\documentclass[conference]{IEEEtran}
\IEEEoverridecommandlockouts

\usepackage{cite}
\usepackage{amsmath,amssymb,amsfonts}
\usepackage{algorithmic}
\usepackage{graphicx}
\usepackage{textcomp}
\usepackage{xcolor}

\usepackage{booktabs}
\usepackage{url}
\usepackage[utf8]{inputenc}

\def\BibTeX{{\rm B\kern-.05em{\sc i\kern-.025em b}\kern-.08em
    T\kern-.1667em\lower.7ex\hbox{E}\kern-.125emX}}
\begin{document}

\title{Personalized Graph-Empowered Large Language Model for Proactive Information Access}

\makeatletter
\newcommand{\linebreakand}{%
  \end{@IEEEauthorhalign}
  \hfill\mbox{}\par
  \mbox{}\hfill\begin{@IEEEauthorhalign}
}
\makeatother

\author{
\IEEEauthorblockN{Chia Cheng Chang}
\IEEEauthorblockA{\textit{Department of Computer Science} \\ \textit{and Information Engineering} \\
\textit{National Taiwan University}\\
Taipei, Taiwan \\
changcc@nlg.csie.ntu.edu.tw}
\and
\IEEEauthorblockN{An-Zi Yen}
\IEEEauthorblockA{\textit{Department of Computer Science} \\
\textit{National Yang Ming Chiao Tung University}\\
Hsinchu, Taiwan \\
azyen@nycu.edu.tw}
\linebreakand
\IEEEauthorblockN{Hen-Hsen Huang}
\IEEEauthorblockA{\textit{Institute of Information Science} \\
\textit{Academia Sinica}\\
Taipei, Taiwan \\
hhhuang@iis.sinica.edu.tw}
\and
\IEEEauthorblockN{Hsin-Hsi Chen}
\IEEEauthorblockA{\textit{Department of Computer Science} \\ \textit{and Information Engineering} \\
\textit{AI Research Center (AINTU)}\\
\textit{National Taiwan University}\\
Taipei, Taiwan \\
hhchen@ntu.edu.tw}
}

\maketitle

\begin{abstract}
Since individuals may struggle to recall all life details and often confuse events, establishing a system to assist users in recalling forgotten experiences is essential.
While numerous studies have proposed memory recall systems, these primarily rely on deep learning techniques that require extensive training and often face data scarcity due to the limited availability of personal lifelogs.
As lifelogs grow over time, systems must also adapt quickly to newly accumulated data.
Recently, large language models (LLMs) have demonstrated remarkable capabilities across various tasks, making them promising for personalized applications.
In this work, we present a framework that leverages LLMs for proactive information access, integrating personal knowledge graphs to enhance the detection of access needs through a refined decision-making process.
Our framework offers high flexibility, enabling the replacement of base models and the modification of fact retrieval methods for continuous improvement.
Experimental results demonstrate that our approach effectively identifies forgotten events, supporting users in recalling past experiences more efficiently.
\end{abstract}


\section{Introduction}

Several works have explored lifelogging from various perspectives. 
\cite{DOHERTY20111948} introduce a method to automatically infer the lifestyle for visual lifelogs. 
\cite{10.5555/2501134.2501176} focus on detecting irregular heart rhythms using a wearable lifelogging camera. 
\cite{6563978} utilize a camera to photo daily meals for weight loss promotion. 
Beyond such applications, several studies explore how past lifelogs can support personal information access services.
In our daily lives, we encounter a variety of events.
As time passes, the details of these past events are often forgotten, leading to potential inaccuracies or omissions when recalling them. 
Hence, a system capable of proactively detecting and assisting in accessing personal information is essential.

\cite{Yen_Huang_Chen_2021} propose a reactive system that can utilize the personal knowledge graphs to answer users' queries about life experience. 
\cite{lin-etal-2022-seen} propose a proactive system to detect four situations where people struggle to recall past experiences in their narratives.
(1)~If the description of a life event matches the user's experience, no memory access assistance is required.
(2)~Since people often forget certain details, they might recall events with similar but unrelated information, leading to conflicts with facts, making it necessary to detect these inconsistencies.
(3)~If certain events are not mentioned, it suggests the user may have forgotten them. The system should remind the user of these forgotten events.
(4)~Users may elaborate on additional events that have not yet been logged. 
In such cases, the system should assist users in updating their personal knowledge graph.

Personal life experiences rapidly accumulate over time, with events continually being updated. 
However, most previous works primarily rely on fine-tuned models~\cite{lin-etal-2022-seen,chu2020multimodal} to support users in recalling experiences from a specific period. 
Consequently, these models face challenges in adapting frequently updated personal knowledge graphs to provide information access service. 
Large language models (LLMs) now serve as the fundamental models for numerous tasks across different domains~\cite{bommasani2021opportunities} by following instructions~\cite{wei2021finetuned,ouyang2022training}.
\cite{lee2024convlogrecaller} conversations focused on memory recall, using LLMs to assist users in recalling experiences with task-specific instructions. When users forget details, the system auto-completes their narratives and identifies discrepancies between the current description and past experiences, offering corrections. 
Preliminary experiments on a limited sample suggest the feasibility of applying LLMs for memory recall assistance.
Therefore, following the work of \cite{lin-etal-2022-seen}, we further investigate the application of LLMs in detecting users' information access needs to address the aforementioned issues. 

Specifically, we propose a framework that comprises three modules.
The base module employs a language model to analyze user descriptions and detect the four situations to provide corresponding services.
The support module integrates structured information from a personal knowledge graph to identify the user's information access needs.
Finally, the correction module synthesizes predictions from the base and support modules. 
Based on our established refinement mechanism, it makes the final decision on which type of service to provide.
We refer to this framework as the Graph-Empowered Refinement (GER) Framework hereafter.
In sum, the contributions of our work are threefold: 
\begin{enumerate}
    \item We explore the utilization of LLMs to support the detection of information access needs. 
    \item A framework is proposed that leverages both textual and structured information to enhance the correctness and reliability of the detection results. 
    \item Experimental results show that our proposed framework achieves promising performance in detecting forgotten life events, and its flexibility allows for the substitution of various base models to enhance performance by incorporating advanced models.
\end{enumerate}

\section{Related Work}

\cite{li-etal-2014-major} propose an approach to extract relevant life events based on tweet descriptions.
Similarly, using social media posts as lifelog data, \cite{10.1145/3331184.3331209} propose a method to extract life events from tweets and construct a personal knowledge base with structured information in the form of subject, predicate, object, and time. 

In addition to analyzing text-based data, \cite{YEN2020102148} investigate the issue of extracting lifelogs from multimodal data.

To assist users in recalling experiences, \cite{10.1145/2911451.2914680,gurrin-2017-overview,gurrin2019overview,10.1145/3372278.3388043} introduce a new task to retrieve the corresponding visual lifelogs with the given query. 
For example, if a user searches for a photo of eating ice cream at the beach, the system should retrieve relevant first-person photos from wearable devices.
Apart from visual lifelogs retrieval, \cite{10.1145/3460426.3463607} categorize the information access service into reactive mode and proactive mode. 
In reactive mode, the user actively initiates a memory recall request, usually by asking about specific details of a life event. 
\cite{Yen_Huang_Chen_2021} develop a question-answering system that can reactively answer past life experiences based on the personal knowledge base with the given question. 
In proactive mode, the system must detect when the user needs memory recall assistance, such as when the user hesitates while recounting an experience~\cite{Wang2018DiscourseMD}, omits details, or provides incorrect event details.
\cite{lin-etal-2022-seen} propose a Structured Event Enhancement Network (SEEN) to support information access service, combining the transformer framework with a graph neural network.
\cite{lee2024convlogrecaller} develop an LLM-based system that detects when users are recalling their experiences during online conversations. 

After the rise of LLMs, many studies have focused on how to leverage graph information within LLMs. 
\cite{10411620} refine raw text using LLMs and extract a knowledge graph (KG) from the refined text via LLMs. 
\cite{10387715} proposes three distinct frameworks to enable LLMs to mutually assist and enhance KGs' capabilities. 
\cite{wen2024mindmap} utilizes KGs to enhance the reasoning capabilities of LLMs and to provide clear visualization of the reasoning process. 
\cite{tang2024graphgpt} integrates the knowledge of LLMs and graphs through graph instruction tuning. 
This work introduces a framework that leverages personal KGs to refine LLM answers and improve detection of information access needs.

\section{Task Definition}


We utilize the NIR dataset~\cite{lin-etal-2022-seen} to investigate the issues of detecting users' information access needs.\footnote{We use only NIR due to the lack of related public datasets.
NIR: \url{https://github.com/ntunlplab/SEEN/tree/main/data/scripts}} 
This dataset comprises two categories of stories: pre-retold story $\mathcal{A}$ and post-retold story $\mathcal{B}$. 
The life experiences were written by users at time point $t$ (for the pre-retold story $\mathcal{A}$) and at time point $t+d$ (for the post-retold story 
$\mathcal{B}$).
Over time, users may forget or misremember events.
\cite{lin-etal-2022-seen} compared these paired stories and annotated five event types. 
Events mentioned in $\mathcal{A}$ need to be checked to ensure they have not been forgotten and omitted in $\mathcal{B}$. 
When an event \textit{e} is mentioned in $\mathcal{B}$, we need to examine whether \textit{e} is consistent with the described experiences previously recorded in $\mathcal{A}$ or if it represents additional new information about past life experiences. 
In NIR, five event types are defined:

\noindent \textbf{Consistent:} The described event in $\mathcal{B}$ matches the user's life experiences in $\mathcal{A}$.
Both stories mention going to the zoo with the user’s girlfriend.

\noindent \textbf{Inconsistent:} The event description in $\mathcal{B}$ conflicts with the facts described in $\mathcal{A}$.
The user mentions ``it was either April or June,'' while $\mathcal{A}$ specifies ``the middle of July,'' showing confusion. 
The system should correct such inaccuracies promptly.

\noindent \textbf{Additional:} Extra information about a life event that was not recorded in $\mathcal{A}$ and does not conflict with the facts.
$\mathcal{B}$ adds new details like ``spiders, snakes, and scorpions,'' which were not in $\mathcal{A}$ but do not conflict with it. 
These details should be stored in the user’s personal KG.

\noindent \textbf{Forgotten:} Life events that have been omitted or not mentioned in $\mathcal{B}$.
The event about ``saw some seals'' is omitted in $\mathcal{B}$, indicating it was forgotten. 
The system should provide reminders.

\noindent \textbf{Unforgotten:} The described event in $\mathcal{A}$ matches the user's life experiences in $\mathcal{B}$.

We extend the setting designed by SEEN~\cite{lin-etal-2022-seen} that uses one story as the reference story and compare event triples in the other story (the target story) with all sentences in the reference story to identify the event type. 
In other words, given a pair of stories, our task is to examine each event triple in the target story and determine its event type by comparing it with all sentences in the reference story.
To identify \textit{Consistent}, \textit{Inconsistent}, and \textit{Additional} events, $\mathcal{A}$ serves as the reference. 
Conversely, to identify \textit{Forgotten} and \textit{Unforgotten} events, $\mathcal{B}$ is used as the reference.
Formally, we define the labels as follows:
\begin{gather*}
y =
\begin{cases} 
\text{Forgotten or Unforgotten} & \text{if } e \in \mathcal{A}, \\
\text{Consistent, Inconsistent, or Additional} & \text{if } e \in \mathcal{B}.
\end{cases}
\end{gather*}
The identified event types correspond to the information access service.
\textit{Unforgotten} and \textit{Consistent} indicate no need for information access.
\textit{Forgotten} means the user has forgotten the life event, so the system needs to provide the recalled event to the user.
\textit{Inconsistent} indicates that the user has mixed up life events, causing the narrative to conflict with the facts. 
The system should remind the user and provide the corrected narrative.
\textit{Additional} represents that the narrative contains extra information about the life event. 
Hence, the system must update the personal KG with the additional event.

\section{Graph-Empowered Refinement Framework}

Our framework comprises three primary components: the base module, the support module, and the correction module. 
The base module is responsible for generating the initial predicted labels through an LLM. 
The support module leverages the knowledge graph and an LLM to identify event triples relevant to the ``query''. 
The ``query'' refers to a sentence composed of event triple (subject, predicate, object) hereafter. For example, a query could be ``I go to my sitting room.'' 
Finally, the correction module refines the labels based on the predictions from both the base and support modules, ultimately determining the final labels.

To sum up, the goal of the three modules is to examine if all the information in the event triples in the target story is relevant to the reference story to determine the initial label.
To support the information access service, after the correction module predicts the final label, we construct a label mapper to convert the relevant or irrelevant labels back to the original event types.

\begin{figure*}[h]
    \centering
    \includegraphics[width=14cm]{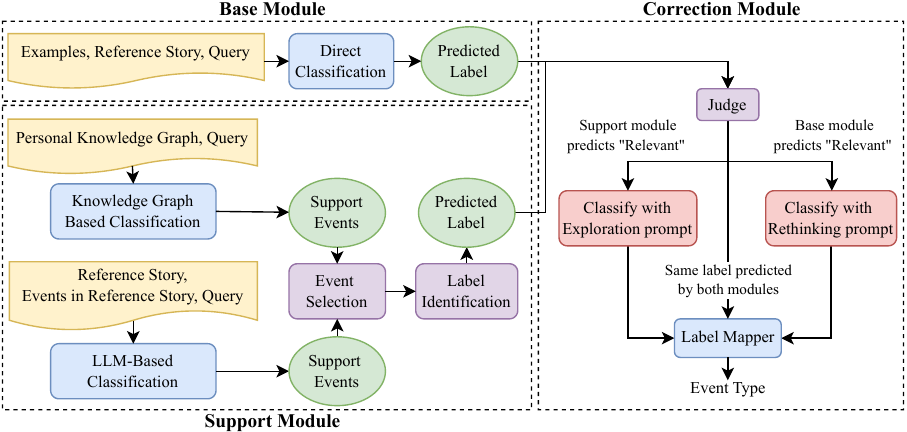}
    \caption{Overview of our graph-empowered refinement (GER) framework for personal information access support.}
    \label{framework}
\end{figure*}

\subsection{Base Module (Direct Prediction)}
The base module is utilized to directly generate initial predictions. 
Given a reference story and a query (i.e., the event triple in the target story), the base module predicts the label. 
Intuitively, it might seem sufficient to use the original labels for classification tasks. 
However, we found that this approach resulted in poor prediction performance.
Consequently, we made adjustments to the labels.

We map \textit{Consistent}, \textit{Inconsistent}, and \textit{Unforgotten} to \textit{Relevant} because the events corresponding to these three types have related descriptions in the reference story. 
In contrast, the events corresponding to \textit{Additional} and \textit{Forgotten} contain information that is completely unrelated to the reference story. 
Thus, we convert these two labels to \textit{Irrelevant}.

\subsection{Support Module (Prediction with Support Event)}
The primary objective of the support module is to identify events in the reference story that are relevant to the query. 
We refer to these events as ``support events''. 
The support module is composed of two components: a knowledge graph-based event classifier and an LLM-based event classifier. 
By using these two classifiers to predict support events, our module can leverage the structured information in the personal knowledge graph and analyze the textual information of each event from the reference story. 
After obtaining the support events predicted by the two classifiers, we determine the final support events by taking the intersection of the two sets.
Since each classifier may identify many irrelevant events, we only consider events identified by both. 
Finally, we determine the label based on whether support events are found. 
If support events are identified, the label is \textit{Relevant}; otherwise, it is \textit{Irrelevant}.

\noindent \textbf{Knowledge Graph-based Classifier.}
In this work, we construct the personal KG by leveraging subjects, predicates, and objects as nodes, and connecting these nodes with edges based on events.
For instance, the event ``me and my girlfriend went to the zoo'' from NIR is represented as (me and my girlfriend, went to, the zoo) as the subject, predicate, and object, respectively. 
Coreference resolution is then applied to link these nodes in the story.
The role of a knowledge graph-based classifier is to identify the support events from the personal KG. 
First, we use the Sentence Transformer~\cite{reimers-gurevych-2019-sentence} to calculate the cosine similarity between the query and nodes in the knowledge graph, where nodes are subjects, predicates, and objects of all the events in KG. 
Nodes with scores above a threshold are identified as key nodes.
Next, we combine the related subject, predicate, and object into event triples. 
Finally, we calculate the score for each triple by combining the scores of the three nodes and select those exceeding the threshold as support events.

\noindent \textbf{LLM-based Classifier.}
To predict support events by LLM, we provide the reference story along with the event from the target story.  
Next, we instruct the LLM to analyze the relationship between each event in the reference story and the query (i.e., the event in the target story).

\subsection{Correction Module (Prediction Correction)}

The correction module refines the labels based on the predicted results from both the base module and the support module using three distinct approaches, each tailored to different cases.
(1)~If both modules' predictions are consistent, the initial label is kept.
(2)~If the base module predicts \textit{Relevant} but the support module predicts \textit{Irrelevant}, the base module may have the issue of hallucination \cite{2311.05232}. 
We apply a rethinking prompt to have LLMs reassess relevance and discard unrelated information. Unlike standard LLM reasoning, which performs an initial inference, the rethinking prompt specifically guides the LLM to reconsider whether the identified relevant information is relevant, reducing hallucinations.
(3)~When the base module predicts \textit{Irrelevant} but the support module predicts \textit{Relevant}, we expect that relevant subgraphs in the KG can capture and supplement details missed by the base module. An exploration prompt is used to guide the LLM in identifying relevant parts based on support events.

\subsection{Label Mapper} \label{label_mapper}
Since our framework initially predicts the relation between the event from the target story and the reference as \textit{Relevant} or \textit{Irrelevant}, we must map the predicted label to the related event type to provide the corresponding information access service.
Hence, a label mapper is constructed.
If the query comes from $\mathcal{A}$, we can directly convert \textit{Relevant} to \textit{Unforgotten} and \textit{Irrelevant} to \textit{Forgotten}.
However, if the query comes from $\mathcal{B}$, the labels (\textit{Consistent}, \textit{Inconsistent}, and \textit{Additional}) are treated as follows: 
If the predicted label is \textit{Irrelevant}, it is directly mapped to \textit{Additional}.
When the predicted label is \textit{Relevant}, it could be either \textit{Consistent} or \textit{Inconsistent}.
In this case, we build a label mapper to further distinguish between \textit{Consistent} and \textit{Inconsistent}.
We employ training instances that are labeled as \textit{Consistent} and \textit{Inconsistent} to fine-tune an LLM to determine the final label. 
The input consists of a query and a reference story. 
The LLM is instructed to follow these steps: 
First, identify the relevant parts in the reference story that relate to the query. 
Second, if the relevant parts do not explicitly contradict the query, label it as \textit{Consistent}. 
Third, if the relevant parts explicitly contradict the query, label it as \textit{Inconsistent}.

\section{Experiments}

\subsection{Baseline Models}
Due to the recent remarkable performance of LLMs in various domains of natural language processing tasks, and their enhanced semantic understanding facilitated by the chain-of-thought \cite{Wei2022ChainOT} approach, we primarily utilize LLMs as the baselines for our task. 
We also compare our models with the state-of-the-art model.  

\noindent \textbf{ChatGPT}: ChatGPT is built upon the generative pre-trained transformer architecture and leverages extensive pre-training on diverse datasets to achieve superior language understanding and generation capabilities. 
A recent study \cite{NEURIPS2020_1457c0d6} has emphasized the effectiveness of LLMs in few-shot learning by utilizing prompt engineering. 
Therefore, we implement a few-shot prompt for ChatGPT in our experiments.  

\noindent \textbf{Llama3} \cite{llama3modelcard}: The architecture of Llama3 leverages advanced transformer mechanisms, enabling it to handle complex linguistic patterns and deliver high-quality, contextually aware outputs. 
Llama3 is particularly noted for its robustness in generating coherent, relevant, and human-like text. 
Considering the model's capabilities and the available resources for our work, we have chosen to use the Llama3-70B as the baseline.

\noindent \textbf{SEEN} \cite{lin-etal-2022-seen}: SEEN is a structured event enhancement network that combines structured information from graphs with textual representations, enabling the model to leverage the strengths of both. 
The comparison with SEEN aims to explore the differences between LLM-based prompting methods and traditional deep learning model training approaches in detecting information access needs.

\subsection{Experiment Setup}
Given that our framework can achieve different enhancement effects by substituting the models within each module, we will introduce the models used in the base module, support module, and correction module. 
In the dataset, there are a total of 6,563 entries, comprising 1,268 \textit{Consistent}, 24 \textit{Inconsistent}, 1,986 \textit{Additional}, 1,897 \textit{Forgotten}, and 1,388 \textit{Unforgotten} cases, respectively.

\noindent \textbf{Base Module.}
In our experiments, we employed ChatGPT, Llama3, SEEN (Base), and SEEN (Large) as the base module. 
As SEEN predicts event types rather than relevant or irrelevant, the predicted ``Consistent'' and ``Unforgotten'' are mapped to ``Relevant'', while the remaining event types are considered ``Irrelevant''.
For ChatGPT, we utilized \texttt{gpt-3.5-turbo-0125}. 
As for Llama3, we employed the \texttt{llama-v3-70b-instruct} model.
The LLM-based baselines are using few-shot prompting.

\noindent \textbf{Support Module.}
In the support module, we categorize our approach into graph-based and LLM-based methods. 
In the graph-based method, we initially employ the ``all-MiniLM-L6-v2''\cite{reimers-2019-sentence-bert} model for encoding the nodes (i.e., subject, predicate, and object) in KG to facilitate subsequent cosine similarity calculations. 
In the LLM-based method, three distinct LLMs were utilized: GPT-3.5, Llama3 70B, and GPT-4o. 

\noindent \textbf{Correction Module.}
In the correction module, we primarily utilize GPT-3.5 for correction. 
Considering resource constraints, we only attempt to further enhance our results by employing GPT-4o in combinations where better outcomes are observed.

\noindent \textbf{Label Mapper.}
To map the predicted \textit{Relevant} and \textit{Irrelevant} labels to the final event types, we fine-tune \texttt{Vicuna-7B}~\cite{vicuna2023} using the following hyperparameters: a learning rate of 2e-5, a sequence length of 768, and one epoch. 

\begin{table}[t]
    \caption{Recall score of each event type.}
    \small
    \centering
    \setlength\tabcolsep{3.8pt}
    \begin{tabular}{lccccc} 
    \toprule
     Model & CST & INC & ADD & FGT & UFG\\
     \hline
     GPT-3.5 & 0.6834 & 0.0000 & 0.7772 & 0.7733 & 0.6961\\ 
     Llama3 70B & 0.7063 & 0.0000 & 0.7846 & 0.7934 & 0.7234\\
     \hline
     SEEN (Base) & 0.7792 & 0.0833 & 0.7905 & 0.8113 & \textbf{0.7500}\\
     SEEN (Large) & \textbf{0.7910} & \textbf{0.1667} & 0.8207 & 0.8371 & 0.7442\\
     \hline
     GER (ours) & 0.7650 & 0.0417 & \textbf{0.8338} & \textbf{0.8635} & 0.7248\\ 
     \bottomrule
    \end{tabular}
    \label{recall-score}
\end{table}

\begin{table}[t]
    \caption{F-score of each event type.}
    \small
    \centering
    \setlength\tabcolsep{3.8pt}
    \begin{tabular}{lccccc} 
     \toprule
     Model & CST & INC & ADD & FGT & UFG \\
     \hline
     GPT-3.5 & 0.6834 & 0.0000 & 0.7772 & 0.7733 & 0.6961 \\ 
     Llama3 70B & 0.7063 & 0.0000 & 0.7846 & 0.7934 & 0.7234 \\
     \hline
     SEEN (Base) & 0.7395 & 0.0952 & 0.8173 & 0.8136 & 0.7470 \\
     SEEN (Large) & \textbf{0.7625} & \textbf{0.2500} & \textbf{0.8374} & 0.8271 & 0.7568 \\
     \hline
     GER & 0.7543 & 0.0741 & 0.8370 & \textbf{0.8364} & \textbf{0.7584} \\ 
     \bottomrule
    \end{tabular}
    \label{f1-result}
\end{table}

\subsection{Experimental Results} \label{experiement-result}
\subsubsection{Main Results}\label{main-results}
The results of the baselines and our framework are shown in Table \ref{recall-score} and Table \ref{f1-result}.  
``Consistent'', ``Inconsistent'', ``Additional'', ``Forgotten'', and ``Unforgotten'' events are abbreviated as CST, INC, ADD, FGT, and UFG, respectively.
``SEEN (Base)'' and ``SEEN (Large)'' refer to the models based on Longformer-base and Longformer-large, respectively.
The primary objective of our task is to detect the need for information access. 
Due to users often forgetting or misremembering details, identifying this need is essential. Thus, the event types \textit{Inconsistent}, \textit{Additional}, and \textit{Forgotten} are considered particularly important for recall services.
To this end, Table~\ref{recall-score} presents the recall scores of detecting five event types.
In Table \ref{recall-score}, our framework with ``SEEN (Large)'' as the base module, GPT-3.5 as the support module for event classification with the graph-based method, and then using GPT-4o to correct the original answers, outperforms all the other baselines in the \textit{Additional} and \textit{Forgotten} event types. 
We measure McNemar's statistical significance test on the baselines and our GRE framework.
The performance of GER in \textit{Forgotten} event type outperforms the baseline models at $p < 0.05$.
This means that our framework allows the correction module to effectively use filtered knowledge graph information to correct answers. Specifically, when life events are forgotten, the support module detects missing relevant information, enabling the correction module to improve predictions and reduce missed reminders for users.
Moreover, we also tested different combinations, and the results show improvement in most \textit{Additional} and \textit{Forgotten} instances compared to ``SEEN (Large)''. 

The F-score of each method is presented in Table~\ref{f1-result}.
Experimental results show that although our framework may not effectively enhance performance in certain event types, it remains comparable to the baselines, and even surpasses them in other event types. 
This indicates that the information provided by the support module from the knowledge graph does contribute to the correction of original answers, leading to better results. 
We will further describe how the information provided by the support module influences the overall outcomes in the following section.
Furthermore, we find that if LLM is used directly for classification, such as the first two rows, the results are inferior compared to other baselines and our framework. 
This also demonstrates the limitations of LLMs in this task, which will be discussed in the following section where we analyze the reasons behind the sub-optimal performance of LLMs. 

\begin{table*}[t]
    \caption{Results with SEEN (Base) as the base module.}
    \small
    \centering
    \begin{tabular}{lccccccc} 
     \toprule
     Base Model & Support Model & Correction Model & Consistent & Inconsistent & Additional & Forgotten & Unforgotten \\
     \hline
     SEEN (Base) & N/A & N/A & 0.7395 & \textbf{0.0952} & 0.8173 & 0.8136 & 0.7470 \\ 
     \hline
     SEEN (Base) & Llama3 70B & GPT-3.5 & 0.7489 & 0.0572 & 0.8261 & 0.8173 & 0.7470 \\
     SEEN (Base) & GPT-3.5 & GPT-3.5 & 0.7401 & 0.0000 & 0.8258 & 0.8212 & 0.7481 \\
     SEEN (Base) & Llama3 70B & GPT-4o & \textbf{0.7516} & 0.0000 & 0.8321 & 0.8243 & \textbf{0.7516} \\
     SEEN (Base) & GPT-3.5 & GPT-4o & 0.7465 & 0.0000 & \textbf{0.8323} & \textbf{0.8376} & 0.7514 \\
     \bottomrule
    \end{tabular}
    \label{base-seen-base}
\end{table*}

\subsubsection{Comparison of Multi-Class and Binary Classification for Event Type Detection}

As previously mentioned, directly instructing an LLM to predict event types yields suboptimal results.
To address this, we transform the task from multi-class classification (i.e., classifying five event types directly) to binary classification (i.e., classifying whether the event is relevant or irrelevant to the reference story).
For the binary classification, the label mapper converts the predicted binary labels into the corresponding event types. 
In Table~\ref{f1-result}, compared to GPT-3.5 and Llama3 70B, which classify event types directly, the GRE method, which utilizes binary classification, is more suitable for LLM-based methods.

\subsubsection{Results with Different Base Module}\label{results-with-different-base}

We investigate whether our framework enhances overall performance even when the base module is replaced. 
Table \ref{base-seen-base} shows the F-score of using ``SEEN (Base)'' as the base module under different settings.
Even though the original results of ``SEEN (Base)'' are significantly better than those obtained by directly using LLMs for classification, our framework can still effectively improve performance across all event types except for \textit{Inconsistent}. 
The overall performance varies across combinations in our framework, leading to improvements across different event types due to the diverse information from the support module.
This also demonstrates that our framework can not only enhance the results of LLMs but also be applied to other models, greatly increasing its flexibility of use.
Furthermore, in the results shown in the last row, it can be observed that even when ``SEEN (Base)'' is used as the base module, our framework effectively corrects wrong predictions, leading to better performance in the \textit{Forgotten} event type compared to ``SEEN (Large)''.

\section{Analysis and Discussion}

\subsection{The Impact of Support Module} \label{impact-support-module}


\begin{table}[t]
\caption{Results of each event type with different support modules. The base module is SEEN (Large), and the correction module is GPT-3.5.}
\begin{center}
\small
\setlength\tabcolsep{4pt}
    \begin{tabular}{lccccc} 
     \toprule
     Model & CST & INC & ADD & FGT & UFG \\
     \hline
     Baseline & 0.7625 & 0.2500 & 0.8374 & 0.8271 & 0.7568 \\ 
     \hline
     Llama3 70B & 0.7576 & 0.0715 & 0.8364 & 0.8237 & 0.7486 \\
     GPT-3.5 & 0.7457 & 0.1480 & 0.8287 & 0.8284 & 0.7495 \\
     Ground Truth & \textbf{0.8546} & \textbf{0.2759} & \textbf{0.9028} & \textbf{0.9014} & \textbf{0.8684} \\
     \bottomrule
    \end{tabular}
\end{center}
\label{diff-support-module}
\end{table}

Our framework, when utilizing ``SEEN (Large)'' as the base module, fails to effectively enhance the performance of certain event types. 
Therefore, we analyze the support module's impact on overall performance.
We replaced the support module with ground truth, providing accurate information to the correction module.
This setup helps us determine if the correction module can rectify more initial classification errors. 
The results in Table~\ref{diff-support-module} show significant improvements across all labels when using ground truth as the support module's output. 
This improvement is greater than with other models, and improving the support module's prediction accuracy enhances overall framework performance.

\subsection{The Importance of Each Module} 

In the previous subsection, we have discussed the importance and the characteristics of the support module.
The capability of the support module determines the upper and lower bounds of the entire framework. 
Hence, we consider the capability of the support module to be the second most important.
In this subsection, we measure the importance of base and correction modules in this framework. 

\begin{table}[t]
\caption{The importance of base module. Both the support module and correction module use GPT-3.5. The model's name indicates the base module used.}
\begin{center}
    \small
    \setlength\tabcolsep{4pt}
    \begin{tabular}{lccccc} 
     \toprule
     Model & CST & INC & ADD & FGT & UFG \\
     \hline
     GPT-3.5 & 0.7163 & 0.0000 & 0.7971 & 0.7914 & 0.7137 \\ 
     Llama3 70B & 0.7204 & 0.0000 & 0.7988 & 0.8004 & 0.7240 \\
     SEEN (Base) & 0.7401 & 0.0000 & 0.8258 & 0.8212 & 0.7481 \\
     SEEN (Large) & \textbf{0.7457} & \textbf{0.1480} & \textbf{0.8287} & \textbf{0.8284} & \textbf{0.7495} \\
     \bottomrule
    \end{tabular}
\end{center}
\label{importance-base-module}
\end{table}

To compare the impact of using different models as our base module, both the support module and the correction module utilize GPT-3.5 as the model. 
The results are shown in Table \ref{importance-base-module}.
``SEEN (Base)'' and ``SEEN (Large)'' denote ``SEEN (Large)'' and ``SEEN (Base)'', respectively.
We find that the performance of the framework is dominated by the capability of the base module. 
In the previous experimental results, the strengths of various baselines are in the following order: ``SEEN (Large)'', ``SEEN (Base)'', Llama3 70B, and GPT-3.5.
Regardless of the event type, the setting of using ``SEEN (Large)'' always achieves the best performance. 
Therefore, we consider the capability of the base module to be the most critical among the three modules. 
When the initial classification results of the base module are suboptimal, even with interventions from the support module and correction module, the overall performance cannot exceed that of a highly capable base module.

\begin{table}[t]
\caption{The importance of the correction module. The first row represents the model of the base module. All the support module uses GPT-3.5.}
\begin{center}
    \small
    \setlength\tabcolsep{3.8pt}
    \begin{tabular}{lccccc} 
     \toprule
     SEEN (Base) & CST & INC & ADD & FGT & UFG \\
     \hline
     GPT-3.5 & 0.7401 & 0.0000 & 0.8258 & 0.8212 & 0.7481 \\ 
     GPT-4o & 0.7465 & 0.0000 & 0.8323 & 0.8376 & 0.7514 \\
     \hline
     \hline
     SEEN (Large) & CST & INC & ADD & FGT & UFG \\
     \hline
     GPT-3.5 & 0.7457 & 0.1480 & 0.8287 & 0.8284 & 0.7495 \\
     GPT-4o & 0.7543 & 0.0741 & 0.8370 & 0.8364 & 0.7584 \\     
     \bottomrule
    \end{tabular}
\end{center}
\label{importance-correction-module}
\end{table}

To measure the impact of the correction module, we compare two different settings: one where the base module was set to ``SEEN (Base)'' and another where it was set to ``SEEN (Large)''. 
The support module is GPT-3.5. 
The results are reported in Table \ref{importance-correction-module}.
The top two rows and the bottom two rows denote the base module are ``SEEN (Base)'' ``SEEN (Large)'', respectively.
The correction module in these two settings is GPT-3.5.
Table \ref{importance-correction-module} shows that regardless of the strength of the base module, when the correction module has a stronger capability, it can further improve the overall performance across most event types. 
The finding suggests that, in our task, the correction module with better semantic understanding capability can more effectively utilize the information provided by the support module.

\subsection{Error Analysis}

\begin{table}[t]
\caption{Prediction analysis of support module with the setting of SEEN (Large), GPT-3.5, GPT-4o.}
\begin{center}
\small
\setlength\tabcolsep{3pt}
    \begin{tabular}{lcccc} 
     \toprule
      & Fail & Success & Alternative Insight & Count \\ 
     \hline
     Consistent & 16.17\% & 58.36\% & 25.47\% & 1,268 \\ 
     Inconsistent & 45.83\% & 20.83\% & 33.34\% & 24 \\
     Additional  & 7.50\% & 74.87\% & 17.63\% & 1,986 \\
     Forgotten & 6.59\% & 77.81\% & 15.60\% & 1,897 \\   
     Unforgotten & 19.24\% & 52.16\% & 28.60\% & 1,388 \\  
     \bottomrule
    \end{tabular}
\end{center}
\label{predict-analysis-support}
\end{table}

We further investigate the contribution of the support module in predicting event types. 
Table \ref{predict-analysis-support} presents the prediction analysis of the support module using the settings of SEEN (Large), GPT-3.5, and GPT-4o.
We decompose the operation of our framework into two parts and analyze the proportion of cases that are correctly or incorrectly predicted. 
This allows us to identify where the primary errors in prediction occur for different event types.

Table \ref{predict-analysis-support} presents the proportions of errors, successes, and alternative insights for each event type, along with their total counts. 
The ``Fail'' column represents that both the base module and support module predict incorrectly. 
The ``Success'' column represents that the base module predicts correctly, and the support module supports the base module's predictions. 
The ``Alternative Insight'' column shows instances where the support module disagrees with the base module, offering alternative suggestions for refining predictions. 
Specifically, if the base module predicts \textit{Relevant}, the support module can suggest that no relevant life events were found. 
Conversely, if the base module predicts \textit{Irrelevant}, the support module can suggest the relevant life events that were found.

We find that our support module performs better in the \textit{Additional} and \textit{Forgotten} event types, supporting the base module's correct predictions in most cases. 
Only 7.5\% of \textit{Additional} cases and 6.59\% of \textit{Forgotten} cases fail to provide alternative insights, indicating no relevant life events in the reference story. This suggests a higher performance ceiling for our framework.
However, for the \textit{Consistent} event type, the support module fails to provide new relevant life events to correct the original erroneous answers in 16.17\% of cases. 
For the \textit{Unforgotten} event type, it fails to find support events in 19.24\% of cases. 
As for the \textit{Inconsistent} event type, there are 45.83\% of error cases. 
Besides the smaller number of cases causing greater variability, the model may interpret misplacement within inconsistent cases as new information, leading to an inability to identify support events.

\begin{table}[t]
\caption{Prediction analysis of correction module with the setting of SEEN (Large), GPT-3.5, GPT-4o.}
\begin{center}
    \small
    \begin{tabular}{lccccc} 
     \toprule
      & Fail & Success & Count\\ 
     \hline
     Consistent & 28.48\% & 71.52\% & 323 \\ 
     Inconsistent & 25.00\% & 75.00\% & 8 \\
     Additional  & 51.71\% & 48.29\% & 350 \\
     Forgotten & 45.27\% & 54.73\% & 296 \\   
     Unforgotten & 28.97\% & 71.03\% & 397 \\  
     \bottomrule
    \end{tabular}
\end{center}
\label{predict-analysis-correction}
\end{table}

Table~\ref{predict-analysis-correction} shows the ratio of successfully corrected cases by the correction module. 
The ``Fail'' and ``Success'' columns represent the failure and success ratios in correcting wrong predictions by our correction module, respectively. 
The results indicate that the correction module performs better in correcting \textit{Consistent}, \textit{Inconsistent}, and \textit{Unforgotten} cases. 
This may be because LLMs tend to reason based on the provided support events, and as long as the support events are not too irrelevant, there is a chance to correct the answers successfully. 
However, the results are suboptimal in the \textit{Additional} and \textit{Forgotten} cases because LLMs sometimes overgeneralize, making it difficult to correct the answers.

\section{Conclusion}
This paper explores the use of LLMs for personal information access support.
A pilot method, the graph-empowered refinement (GER) framework, is proposed. 
GER enables LLMs to leverage the information contained within personal knowledge graphs and refine the generated responses based on the provided information. 
The experimental results also demonstrate that different models can be applied within our framework, leading to further improvements. 
This flexibility in our framework allows it to be utilized not only with specific models but also with advanced models in the future, thereby enhancing performance through our framework. 
Furthermore, unlike traditional  models, our approach does not require training. 
This means our method can quickly and effectively adapt to users' continuously accumulating life events over time, providing support for information access services.

As indicated by our experiments, directly using LLMs for classification leads to suboptimal performance. 
Through case studies, we can observe that for the \textit{Additional} and \textit{Forgotten} event types, LLMs often exhibit hallucinations, leading them to incorrectly perceive the query as relevant to the story. 
Additionally, our definitions for \textit{Additional} and \textit{Forgotten} are relatively strict, but LLMs struggle to accurately discern whether a sentence contains entirely new information. 
This also causes them to incorrectly identify queries as matching previous information or not being forgotten. 
For \textit{Consistent}, \textit{Inconsistent}, and \textit{Unforgotten}, LLMs may fail to extract relevant parts due to a change in the description of the same event within the story, resulting in the inability to correctly identify them. 
Another key limitation is the handling of user privacy when building and using personal knowledge graphs. 
Developing a framework to ensure data security is left as future work.
In the future, we will establish an end-to-end system to practically utilize in daily life for extracting life events and provide enhanced information access support, such as correcting conflicting descriptions.

\section*{Acknowledgment}
This research was partially supported by Google gift toward the work on Multimodal User Profiling for Personalized Information Access (Reference Id 00030586), National Science and Technology Council, Taiwan, under grants NSTC STC 114-2221-E-002-070-MY3 and NSTC 113-2634-F-002-003-, and Ministry of Education (MOE) in Taiwan, under grants NTU-114L900901.


\bibliographystyle{IEEEtran}
\bibliography{IEEEabrv}

\end{document}